\documentclass{article} 
\usepackage{iclr2016_conference,times}
\usepackage{hyperref}
\usepackage{url}
\usepackage[pdftex]{graphicx}

\title{ LSTM-based Deep Learning Models for non-factoid answer selection}

\author{Ming Tan, Cicero dos Santos, Bing Xiang \& Bowen Zhou\\
IBM Watson Core Technologies\\
Yorktown Heights, NY, USA \\
\texttt{\{mingtan,cicerons,bingxia,zhou\}@us.ibm.com} \\
}

\begin{document}

\maketitle

\begin{abstract}

In this paper, we apply a general deep learning (DL) framework for the answer selection task, which does not depend on manually defined features or linguistic tools. The basic framework is to build the embeddings of questions and answers based on bidirectional long short-term memory (biLSTM) models, and measure their closeness by cosine similarity. We further extend this basic model in two directions. One direction is to define a more composite representation for questions and answers by combining convolutional neural network with the basic framework. The other direction is to utilize a simple but efficient attention mechanism in order to generate the answer representation according to the question context. Several variations of models are provided. The models are examined by two datasets, including TREC-QA and InsuranceQA. Experimental results demonstrate that the proposed models substantially outperform several strong baselines.


\end{abstract}

\section{Introduction}

The answer selection problem can be formulated as follows: Given a question $q$ and an answer candidate pool
$\{a_1, a_2, \cdots , a_s\}$ for this question, we aim to search for the best answer candidate $a_k$, where $1 \le k \le s$. An answer is a token sequence with an arbitrary length, and a question can correspond to multiple ground-truth answers. In testing, the candidate answers for a question may not be observed in the training phase. Answer selection is one of the essential components in typical question answering (QA) systems. It is also a stand-alone task with applications in knowledge base construction and information extraction.

The major challenge of this task is that the correct answer might not directly share lexical units with the question. Instead, they may only be semantically related. Moreover, the answers are sometimes noisy and contain a large amount of unrelated information.

Recently, deep learning models have obtained a significant success on various natural language processing tasks, such as semantic analysis \citep{tang2015}, machine translation \citep{bahdanau2015} and text summarization \citep{rush2015}. 

In this paper, we propose a deep learning framework for answer selection which does not require any feature engineering, linguistic tools, or external resources. This framework is based on building bi-directional long short term memory (biLSTM) models on both questions and answers respectively, connecting with a pooling layer and utilizing a similarity metric to measure the matching degree. We improve this basic model from two perspectives. Firstly, a simple pooling layer may suffer from the incapability of keeping the local linguistic information. In order to obtain better embeddings for the questions and answers, we build a convolutional neural network (CNN) structure on top of biLSTM. Secondly, in order to better distinguish candidate answers according to the question, we introduce a simple but efficient attention model to this framework for the answer embedding generation according to the question context.

We report experimental results for two answer selection datasets: (1) InsuranceQA \citep{minwei15} \footnote{git clone https://github.com/shuzi/insuranceQA.git}, 
a recently released large-scale non-factoid QA dataset from the insurance domain. The proposed models demonstrate a significant out-performance compared to two non-DL baselines and a strong DL baseline based on CNN. (2) TREC-QA \footnote{The data is obtained from \citep{yao2013} \url{http://cs.jhu.edu/~xuchen/packages/jacana-qa-naacl2013-data-results.tar.bz2}},  which was created by \cite{wangmengqiu2007} based on Text REtrieval Conference (TREC) QA track data. The proposed models 
outperform various strong baselines.

The rest of the paper is organized as follows: Section 2 describes the related work for answer selection; Section 3 provides the details of the proposed models; Experimental settings and results of InsuranceQA and TREC-QA datasets are discussed in section 4 and 5 respectively; Finally, we draw conclusions in section 6.

\section{Related work}

Previous work on answer selection normally used feature engineering, linguistic tools, or external resources. For example, semantic features were constructed based on WordNet in \citep{yih2013}. This model pairs semantically related words based on word semantic relations.
In \citep{wangmengqiu2010,wangmengqiu2007}, the answer selection problem is transformed to a syntactical matching between the question/answer parse trees. Some work tried to fulfill the matching using minimal edit sequences between dependency parse trees \citep{heilman2010,yao2013}. Recently, discriminative tree-edit features extraction and engineering over parsing trees were automated in \citep{severyn2013}. 

While these methods show effectiveness, they might suffer from the availability of additional resources, the effort of feature engineering and the systematic complexity by introducing linguistic tools, such as parse trees and dependency trees.  

There were prior methods using deep learning technologies for the answer selection task. The approaches for non-factoid question answering generally pursue the solution on the following directions: Firstly, the question and answer representations are learned and matched by certain similarity metrics \citep{minwei15,yu2014,dossantos2015}.  Secondly, a joint feature vector is constructed based on both the question and the answer, and then the task can be converted into a classification or learning-to-rank problem \citep{wang2015}. Finally, recently proposed models for textual generation can intrinsically be used for answer selection and generation \citep{bahdanau2015,oriol2015}. 

The framework proposed in this work belongs to the first category. There are two major differences between our approaches and the work in \citep{minwei15}: (1) The architectures developed in \citep{minwei15} are only based on CNN, whereas our models are based on bidirectional LSTMs, which are more capable of exploiting long-range sequential context information. Moreover, we also integrate the CNN structures on the top of biLSTM for better performance. (2) \cite{minwei15} tackle the question and answer independently, while the proposed structures develop an efficient attentive models to generate answer embeddings according to the question. 

\section{Approach}

In this section, we describe the proposed framework and its variations. We first introduce the general framework, which is to build bi-directional LSTM on both questions and their answer candidates, and then use the similarity metric to measure the distance of question answer pairs. In the following two subsections, we extend the basic model in two independent directions.

\subsection{Basic Model: QA-LSTM} 

{\bf Long Short-Term Memory (LSTM): } Recurrent Neural Networks (RNN) have been widely exploited to deal with variable-length sequence input. The long-distance history is stored in a recurrent hidden vector which is dependent on the immediate previous hidden vector. LSTM \citep{lstm1997} is one of the popular variations of RNN to mitigate the gradient vanish problem of RNN. Our LSTM implementation is similar to the one in \citep{graves2013} with minor modification. Given an input sequence $\mathbf{x}=\{\mathbf{x}(1), \mathbf{x}(2), \cdots, \mathbf{x}(n) \}$, where $\mathbf{x}(t)$ is an $E$-dimension word vector in this paper. The hidden vector $\mathbf{h}(t)$ ( the size is $H$ ) at the time step $t$ is updated as follows.  

\begin{eqnarray}
i_{t} & = & \sigma(\mathbf{W}_{i}\mathbf{x}(t)+\mathbf{U}_{i}\mathbf{h}(t-1)+\mathbf{b}_{i})\\
f_{t} & = & \sigma(\mathbf{W}_{f}\mathbf{x}(t)+\mathbf{U}_{f}\mathbf{h}(t-1)+\mathbf{b}_{f})\\
o_{t} & = & \sigma(\mathbf{W}_{o}\mathbf{x}(t)+\mathbf{U}_{o}\mathbf{h}(t-1)+\mathbf{b}_{o})\\
\tilde{C}_{t} & = & \tanh(\mathbf{W}_{c}\mathbf{x}(t)+\mathbf{U}_{c}\mathbf{h}(t-1)+\mathbf{b}_{c})\\
C_{t} & = & i_{t}*\tilde{C}_{t}+f_{t}*C_{t-1}\\
\mathbf{h}_{t} & = & o_{t}*\tanh(C_{t})
\end{eqnarray}

In the LSTM architecture, there are three gates
(input $i$, forget $f$ and output $o$), and a cell memory vector $c$. 
$\sigma$ is the $sigmoid$ function. The input gate can determine how incoming vectors $x_t$ alter the state of the memory cell. The output gate can allow the memory cell to have an effect on the outputs. Finally, the forget gate allows the cell to remember or forget its previous state.
$\mathbf{W} \in R^{H \times E}$, $\mathbf{U} \in R^{H \times H}$ and $\mathbf{b} \in R^{H \times 1}$ are the network parameters. 

{\bf Bidirectional Long Short-Term Memory (biLSTM): } Single direction LSTMs suffer a weakness of not utilizing the contextual information from the future tokens. Bidirectional LSTM utilizes both the previous and future context by processing the sequence on two directions, and generate two independent sequences of LSTM output vectors. One processes the input sequence in the forward direction, while
the other processes the input in the reverse direction.
The output at each time step is the concatenation of the two output vectors from both directions, ie. $h_t$ = $\overrightarrow{h_{t}} \parallel \overleftarrow{h_{t}}$.

\begin{figure}[b]
    \centering
    \includegraphics[width=0.5\textwidth]{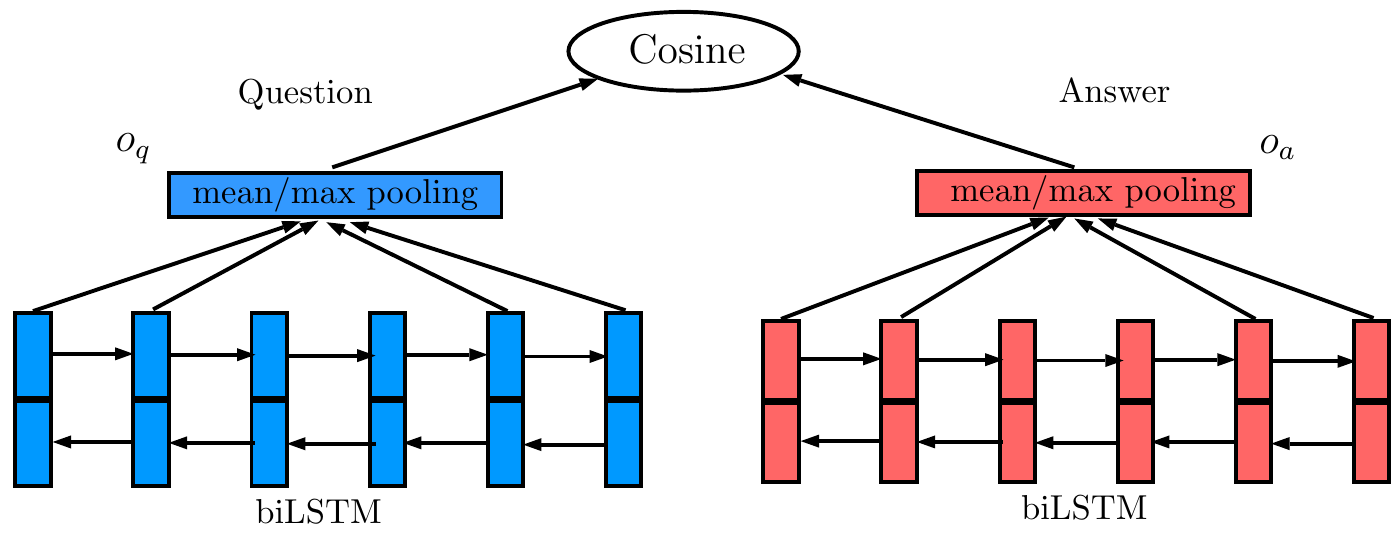}
    \caption{Basic Model: QA-LSTM}
    \label{fig:qalstm}
\end{figure}

{\bf QA-LSTM:} The basic model in this work is shown in Figure \ref{fig:qalstm}. BiLSTM generates distributed representations for both the question and answer independently, and then utilize cosine similarity to measure their distance. Following the same ranking loss in \citep{minwei15,weston2014,hu2014}, we define the training objective as a hinge loss. 
\begin{equation}
L = \max \{ 0, M-cosine(q, a_+)+cosine(q, a_-) \}
\end{equation}
where $a_+$ is a ground truth answer, $a_-$ is an incorrect answer randomly chosen from the entire answer space, and $M$ is constant margin. We treat any question with more than one ground truth as multiple training examples, each for one ground truth. 

There are three simple ways to generate representations for questions and answers based on the word-level biLSTM outputs: (1) Average pooling; (2) max pooling; (3) the concatenation of the last vectors on both directions. The three strategies are compared with the experimental performance in Section 5. Dropout operation is performed on the QA representations before cosine similarity matching. 

Finally, from preliminary experiments, we observe that the architectures, in which both question and answer sides share the same network parameters, is significantly better than the one that the question and answer sides own their own parameters separately, and converges much faster. As discussed in \citep{minwei15}, this is reasonable, because for a shared layer network, the corresponding elements in question and answer vectors represent the same biLSTM outputs. While for the network with separate question and answer parameters, there is no such constraint and the model has double-sized parameters, making it difficult to learn for the optimizer. 

\subsection{QA-LSTM/CNN}

In the previous subsection, we generate the question and answer representations only by simple operations, such as max or mean pooling.
In this subsection, we resort to a CNN structure built on the outputs of biLSTM, in order to give a more composite representation of questions and answers. 

\begin{figure}[t]
    \centering
    \includegraphics[width=0.55\textwidth]{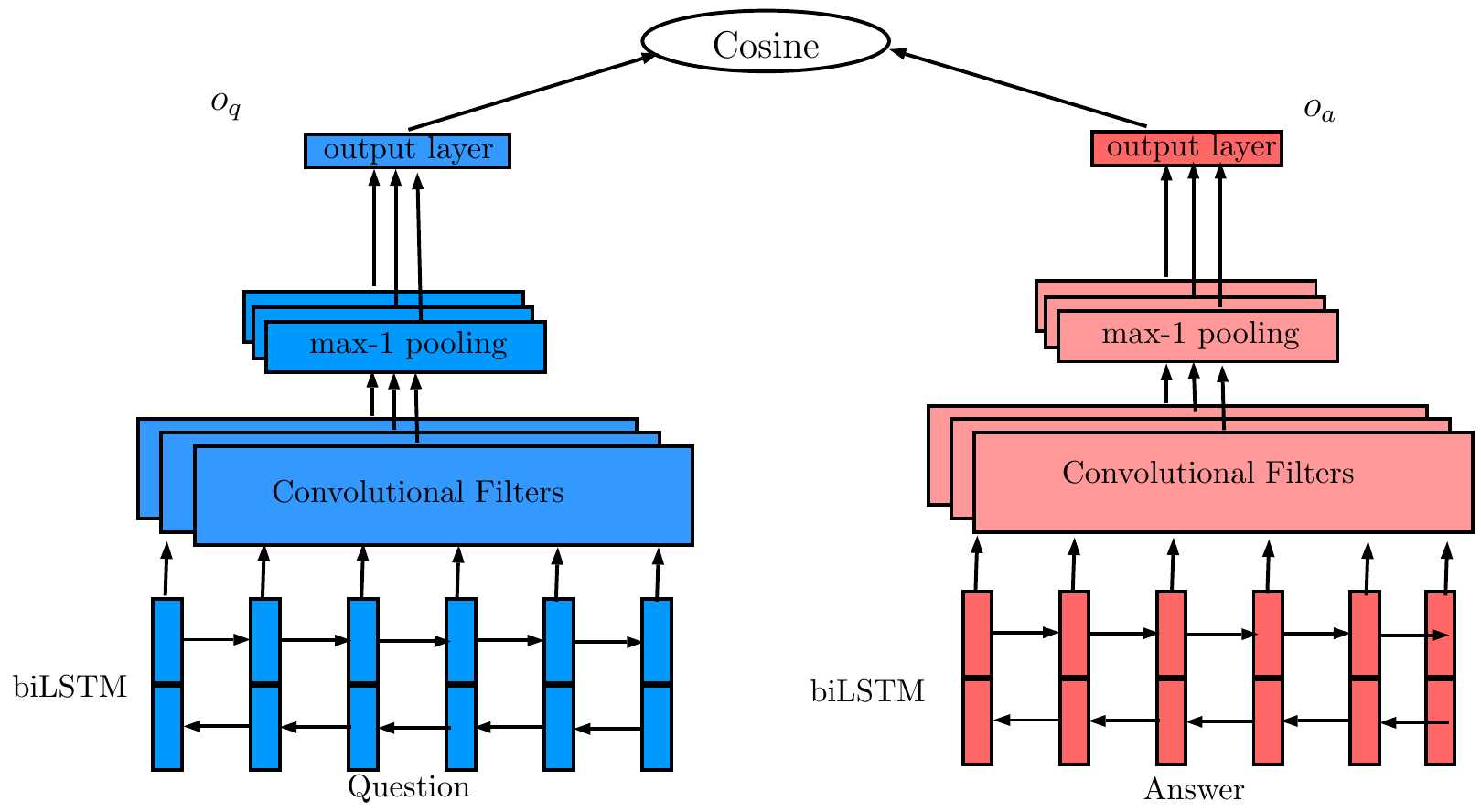}
    \caption{QA-LSTM/CNN}
    \label{fig:qalstmcnn}
\end{figure}

The structure of CNN in this work is similar to the one in \citep{minwei15}, as shown in Figure \ref{fig:qalstmcnn}. Unlike the traditional forward neural network, where each output is interactive with each input, the convolutional structure only imposes local interactions between the inputs within a filter size $m$. 

In this work, for every window with the size of $m$ in biLSTM output vectors, ie. $\mathbf{H}_m(t)=[\mathbf{h}(t), \mathbf{h}(t+1), \cdots, \mathbf{h}(t+m-1)]$, where $t$ is a certain time step, the convolutional filter $\mathbf{F}=[\mathbf{F}(0) \cdots \mathbf{F}(m-1)]$ will generate one value as follows. 

\begin{eqnarray}
o_{F}(t) & = & \tanh\left[\left(\sum_{i=0}^{m-1}\mathbf{h}(t+i)^{T}\mathbf{F}(i)\right){\displaystyle +b}\right]
\end{eqnarray}
where $b$ is a bias, and $\mathbf{F}$ and $b$ are the parameters of this single filter.

Same as typical CNNs, a max-$k$ pooling layer is built on the top of the convolutional layer. Intuitively, we want to emphasize the top-$k$ values from each convolutional filter. By $k$-MaxPooling, the maximum $k$ values will be kept for one filter, which indicate the highest degree that a filter matches the input sequence.

Finally, there are $N$ parallel filters, with different parameter initialization, and the convolutional layer
gets $N$-dimension output vectors. We get two output vectors with dimension of $kN$ for the questions and answers respectively. In this work, $k=1$. $k>1$ did not show any obvious improvement in our early experiments. The intuition of this structure is, instead of evenly considering the lexical information of each token as the previous subsection, we emphasize on certain parts of the answer, such that QA-LSTM/CNN can more effectively differentiate the ground truths and incorrect answers. 

\subsection{Attention-based QA-LSTM }

In the previous subsection, we described one extension from the basic model, which targets at providing more composite embeddings for questions and answers respectively. In this subsection, we investigate an extension from another perspective. Instead of generating QA representation independently, we leverage a simple attention model for the answer vector generation based on questions. 

The fixed width of hidden vectors becomes a bottleneck, when the bidirectional LSTM models must propagate dependencies over long distances over the questions and answers. An attention mechanism are used to alleviate this weakness by dynamically aligning the more informative parts of answers to the questions. This strategy has been used in many other natural language processing tasks, such as machine translation \citep{bahdanau2015,sutskever2014}, sentence summarization \citep{rush2015} and factoid question answering \citep{herman2015,sukhbaatar2015}. 
\begin{figure}[b]
    \centering
    \includegraphics[width=0.55\textwidth]{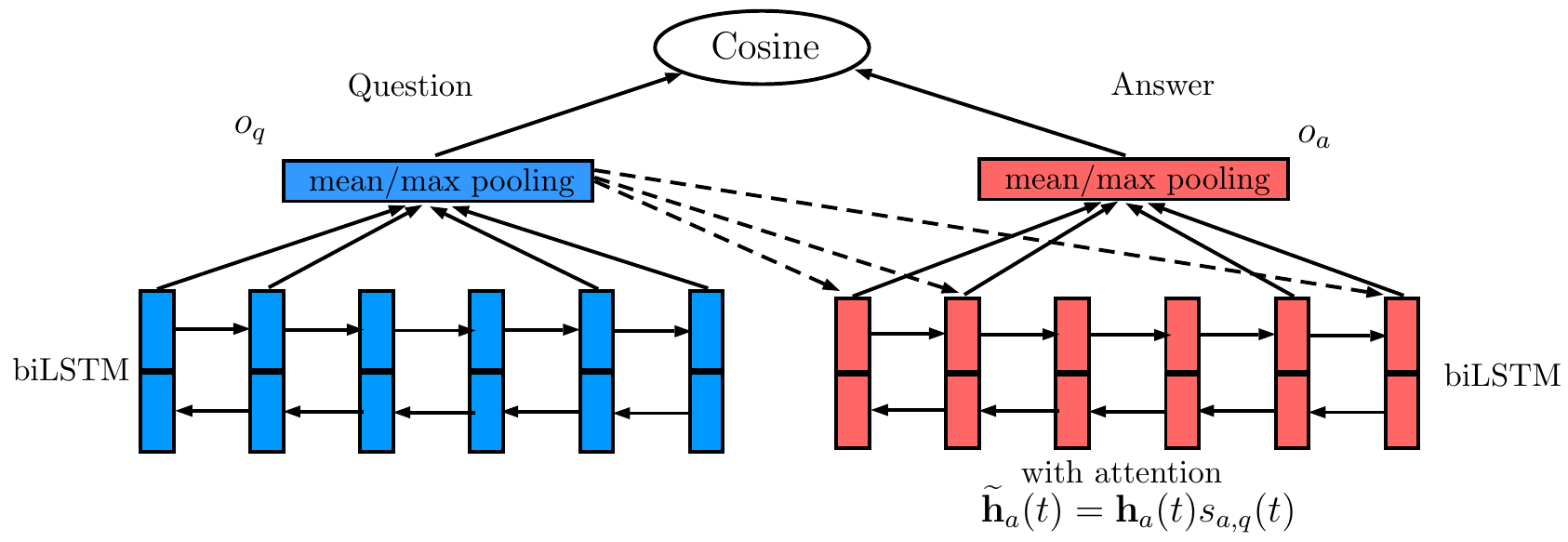}
    \caption{QA-LSTM with attention}
    \label{fig:qalstmatt}
\end{figure}

Inspired by the work in \citep{herman2015}, we develop a very simple but efficient word-level attention on the basic model. Figure \ref{fig:qalstmatt} shows the structure. Prior to the average or mean pooling, each biLSTM output vector will be multiplied by a softmax weight, which is determined by the question embedding from biLSTM. 

Specifically, given the output vector of biLSTM on the answer side at time step $t$, $\mathbf{h}_{a}(t)$, and the question embedding, $\mathbf{o}_q$, the updated vector $\mathbf{\widetilde{h}}_{a}(t)$ for each answer token are formulated below. 

\begin{eqnarray}
\mathbf{m}_{a,q}(t) & = & \tanh(\mathbf{W}_{am}\mathbf{h}_{a}(t)+\mathbf{W}_{qm}\mathbf{o}_{q})\\
s_{a,q}(t) & \propto &  \exp(\mathbf{w}_{ms}^{T}\mathbf{m}_{a,q}(t))\\
\mathbf{\widetilde{h}}_{a}(t) & = & \mathbf{h}_{a}(t)s_{a,q}(t)
\end{eqnarray}
where $\mathbf{W}_{am}$, $\mathbf{W}_{qm}$ and $\mathbf{w}_{ms}$ are attention parameters. 
Conceptually, the attention mechanism give more weights on certain words, just like tf-idf for each word. However, the former computes the weights according to question information.

The major difference between this approach and the one in \citep{herman2015} is that \cite{herman2015}'s attentive reader emphasizes the informative part of supporting facts, and then uses a combined embedding of the query and the supporting facts to predict the factoid answers. In this work, we directly use the attention-based representations to measure the question/answer distances. Experiments indicate the attention mechanism can more efficiently distinguish correct answers from incorrect ones according to the question text. 

\subsection{QA-LSTM/CNN with attention}

The two extensions introduced previously are combined in a simple manner. First, the biLSTM hidden vectors of answers $\mathbf{h}_{a}(t)$ are multiplied by $s_{a,q}(t)$, which is computed from the question average pooling vectors $o_q$, and updated to $\mathbf{\widetilde{h}}_{a}(t)$, illustrated in Eq. 9-11. Then, the original question and updated answer hidden vectors serve as inputs of CNN structure respectively, such that the question context can be used to evaluate the softmax weights of the input of CNN. From the experiments, we observe that the two extensions vary on their contributions on the performance improvement according to different datasets. However, QA-LSTM/CNN with attention can outperform the baselines on both datasets.

\begin{table}
\begin{center}
\begin{tabular}{|l|llll|}
\hline
 & \bf Train  & \bf Validation & \bf Test1 & \bf Test2 \\
\hline
\# of Qs & 12887 & 1000 & 1800 & 1800 \\
\# of As & 18540 & 1454 &  2616 & 2593 \\
\hline
\end{tabular}
\end{center}
\caption{\label{numberq} Numbers of questions and answers of InsuranceQA. }
\vspace{-2mm}
\end{table}

\section{InsuranceQA Experiments}
\vspace{-2mm}
Having described a number of models in the previous section, we evaluate the proposed approaches on the insurance domain dataset, InsuranceQA, provided by \cite{minwei15}. The InsuranceQA dataset provides a training set, a validation set, and two test sets. We do not see obvious categorical differentiation between two tests' questions. One can see the details of InsuranceQA data in \citep{minwei15}. We list the numbers of questions and answers of the dataset in Table \ref{numberq}. A question may correspond to multiple answers. The questions are much shorter than answers. The average length of questions is 7, and the average length of answers is 94. The long answers comparing to the questions post challenges for answer selection task. This corpus contains 24981 unique answers in total. For the development and test sets, the dataset also includes an answer pool of 500 candidate answers for each question. These answer pools were constructed by including the correct answer(s) and randomly selecting candidate from the complete set of unique answers. The top-1 accuracy of the answer pool is reported.

\subsection{Setup}

The models in this work are implemented with Theano \citep{Bastien-Theano-2012} from scratch, and all experiments are processed in a GPU cluster.  We use the accuracy on validation set to locate the best epoch and best hyper-parameter settings for testing.

The word embedding is trained by word2vec \citep{word2vec2013}, and the word vector size is 100. Word embeddings are also parameters and are optimized as well during the training. Stochastic Gradient Descent (SGD) is the optimization strategy. 
We tried different margin values, such as 0.05, 0.1 and 0.2, and finally fixed the margin as 0.2. We also tried to include $l_2$ norm in the training objective. However, preliminary experiments show that regularization factors do not show any improvements. Also, the dimension of LSTM output vectors is 141 for one direction, such that biLSTM has a comparable number of parameters with a single-direction LSTM with 200 dimension.

We train our models in mini-batches (the batch size $B$ is 20), and the maximum length $L$ of questions and answers is 200. Any tokens out of this range will be discarded. Because the questions or answers within a mini-batch may have different lengths, we resort to a mask matrix $M \in R^{B \times L} $ to indicate the real length of each token sequence. 

\subsection{Baselines}
For comparison, we report the performances of four baselines in Table \ref{baselines}: two state-of-the-art non-DL approaches and two variations of a strong DL approach based on CNN as follows.

\begin{table*}
\begin{center}
\begin{tabular}{|l|lll|}
\hline
 & \bf Validation  & \bf Test1 & \bf Test2 \\
\hline
A. Bag-of-word & 31.9 & 32.1 & 32.2 \\
B. Metzler-Bendersky IR model & 52.7 & 55.1 &  50.8 \\
C. Architecture-II in \citep{minwei15} & 61.8 & 62.8 & 59.2  \\
D. Architecture-II with GESD & {\bf 65.4} & {\bf 65.3} & {\bf 61.0}  \\
\hline
\end{tabular}
\end{center}
\caption{\label{baselines} Baseline results of InsuranceQA }
\vspace{-2mm}
\end{table*}

{\bf Bag-of-word}: The idf-weighted sum of word vectors for the question and for all of its answer candidates is used as a feature vector. Similar to this work, the candidates are re-ranked according the cosine similarity to a question.

{\bf Metzler-Bendersky IR model}: A state-of-the-art weighted dependency (WD) model, which employs a weighted combination of term-based and term proximity-based ranking features to score each candidate answer.

\begin{table*}[t]
\begin{center}
\begin{tabular}{|l|llll|}
 \hline 
& \bf Model & \bf Validation  & \bf Test1 & \bf Test2 \\
 \hline 
A & QA-LSTM basic-model(head/tail) & 54.0 & 53.1 & 51.2 \\
B & QA-LSTM basic-model(avg pooling) & 58.5 & 58.2 & 54.0 \\
C & QA-LSTM basic-model(max pooling) & 64.3 & 63.1 & 58.0 \\
\hline
D & QA-LSTM/CNN(fcount=1000) & 65.5 & 65.9 & 62.3 \\
E & QA-LSTM/CNN(fcount=2000) & 64.8 & 66.8 & 62.6 \\
F & QA-LSTM/CNN(fcount=4000) & 66.2 & 64.6 & 62.2  \\
\hline
G & QA-LSTM with attention (max pooling) & 66.5 & 63.7 & 60.3  \\
H & QA-LSTM with attention (avg pooling) & {\bf 68.4} & {\bf 68.1} & 62.2 \\
I & QA-LSTM/CNN (fcount=4000) with attention & 67.2 & 65.7 & {\bf 63.3} \\
 \hline 
\end{tabular}
\end{center}
\caption{\label{results} The experimental results of InsuranceQA for QA-LSTM, QA-LSTM/CNN and QA-LSTM with attentions }
\vspace{-2mm}
\end{table*}

{\bf Architecture-II in \citep{minwei15}}: Instead of using LSTM, a CNN model is employed to learn a distributed vector representation of a given question and its answer candidates, and the answers are scored by cosine similarity with the question. No attention model is used in this baseline. 

{\bf Architecture-II with Geometricmean of Euclidean and Sigmoid Dot product (GESD)}: GESD is used to measure the distance between the question and answers. This is the model which achieved the best performance in \citep{minwei15}.

\subsection{Results and discussions}

In this section, detailed analysis on experimental results are
given. Table \ref{results} summarizes the results of our models on InsuranceQA. 
From Row (A) to (C), we list QA-LSTM without either CNN structure or attention model. They vary on how to utilize the biLSTM output vectors to form sentential embeddings for questions and answers in shown in section 3.1. We can observe that just concatenating of the last vectors from both direction (A) performs the worst. It is surprised to see using max-pooling (C) is much better than average pooling (B). The potential reason is that the max-pooling extracts more local values for each dimension, so that more local information can be reflected on the output embeddings.

From Row (D) to (F), CNN layers are built on the top of the biLSTM with different filter numbers. We set the filter width $m=2$, and we did not see better performance if we increase $m$ to 3 or 4.  
Row (F) with 4000 filters gets the best validation accuracy, obtained a comparable performance with the best baseline (Row (D) in Table \ref{baselines} ).  Row F shared a highly analogous CNN structure with Architecture II in \citep{minwei15}, except that the later used a shallow hidden layer to transform the word embeddings into the input of CNN structure, while Row F take the output of biLSTM as CNN input. 

Row (G) and (H) corresponds to QA-LSTM with the attention model. (G) connects the output vectors of answers after attention with a max pooling layer, and (H) with an average pooling. In comparison to Model (C), Model (G) shows over 2\% improvement on both validation and Test2 sets. With respect to the model with mean pooling layers (B), the improvement from attention is more remarkable. Model (H) is over 8\% higher on all datasets compared to (B), and gets improvements from the best baseline by 3\%, 2.8\% and 1.2\% on the validation, Test1 and Test2 sets, respectively. Compared to Architecture II in \citep{minwei15}, which involved a large number of CNN filters, (H) model also has fewer parameters. 

Row (I) corresponds to section 3.4, where CNN and attention mechanism are combined. Although compared to (F), it shows 1\% improvement on all sets, we fail to see obvious improvements compared to Model (H). Although Model (I) achieves better number on Test2, but does not on validation and Test1. We assume that the effective attention might have vanished during the CNN operations. However, both (H) and (I) outperform all baselines.

\begin{figure*}[t]
\begin{centering}
\includegraphics[scale=0.32]{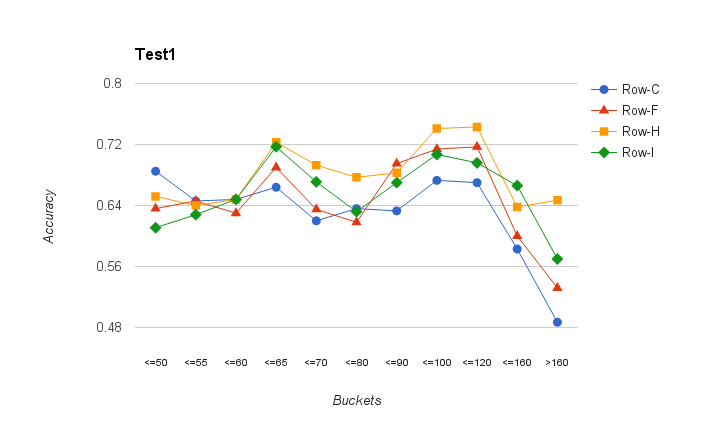}\includegraphics[scale=0.32]{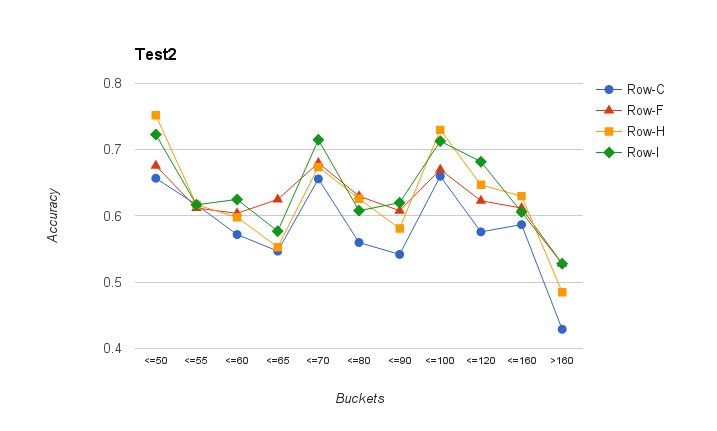}

\begin{tabular}{|l|lllllllllll|}
\hline 
\bf Buckets & \bf $\le$50 & \bf $\le$55 & \bf $\le$60 & \bf $\le$65 & \bf $\le$70 & \bf $\le$80 & \bf $\le$90 & \bf $\le$100 & \bf $\le$120 & \bf $\le$160 & \bf $>$160 \tabularnewline
\hline
Test1 & 121 & 167 & 165 & 152 & 137 & 223 & 161 & 147 & 191 & 180 & 156 \tabularnewline
Test2 & 105 & 191 & 192 & 168 & 169 & 230 & 153 & 115 & 170 & 165 & 142 \tabularnewline
\hline
\end{tabular}
\par

\par\end{centering}
\caption{The accuracy of Test1 and Test2 of InsuranceQA sets for the four models (C, H, F and I in Table \ref{results} ), on different levels of ground truth answer lengths. The table divided each test set into 11 buckets. The figures above show the accuracy of each bucket. }
\label{fig:testbucket}
\vspace{-2mm}
\end{figure*}

We also investigate the proposed models on how they perform with respect to long answers. We divide the questions of Test1 and Test2 sets into eleven buckets, according to the average length of their ground truths. In the table of Figure \ref{fig:testbucket}, we list the bucket levels and the number of questions which belong to each bucket, for example, Test1 has 165 questions, whose average ground truth lengths are $55 < L \le 60$. We select models of (C), (F), (H) and (I) in Table \ref{results} for comparison. Model (C) is without attention and sentential embeddings are formed only by max pooling. Model (F) utilizes CNN, while model (H) and (I) integrate attention. As shown in the left figure in Figure \ref{fig:testbucket}, (C) gets better or close performance compared to other models on buckets with shorter answers ($\le$ 50, $\le$55, $\le$60). However, as the ground lengths increase, the gap between (C) and other models becomes more obvious. The similar phenomenon is also observed in the right figure for Test2. This suggests the effectiveness of the two extensions from the basic model of QA-LSTM, especially for long-answer questions.

\cite{minwei15} report that GESD outperforms cosine similarity in their models. However, the proposed models with GESD as similarity scores do not provide any improvement on accuracy. 

Finally, we replace the cosine similarity with a MLP structure, whose input (282x2-dimension) is the concatenation of question and answer embeddings, and the output is a single similarity score and test the modified models by a variety of hidden layer size (100,500,1000). We observe that the modified models not only get $>$10\% accuracy decrease, but also converge much slower. One possible explanation is the involvement of more network parameters by MLP makes it more difficult for training, although we believed that MLP might partially avoid the conceptual challenge of projecting questions and answers in the same high-dimensional space, introduced by cosine similarity. 
\vspace{-1mm}
\section{TREC-QA Experiments}
\vspace{-2mm}
In this section we detail our experimental setup and results using the TREC-QA dataset.

\begin{table}
\small
\centering
\begin{tabular}{|l|ll|}
\hline  
\bf Models & \bf MAP & \bf MRR \\
\hline
\cite{wangmengqiu2007} & 0.6029 & 0.6852 \\
\cite{heilman2010} & 0.6091 & 0.6917 \\
\cite{wangmengqiu2010} & 0.6029 & 0.6852 \\
\cite{yao2013} & 0.6307 & 0.7477 \\
\cite{severyn2013} & 0.6781 & 0.7358\\
\cite{yih2013}-BDT & 0.6940 & 0.7894\\
\cite{yih2013}-LCLR & 0.7092 & 0.7700 \\
\cite{wang2015} & \bf 0.7134 & 0.7913 \\
Architecture-II \citep{minwei15} & 0.7106 & \bf 0.7998 \\
\hline 
\end{tabular}
\caption{\label{trec-qa-baselines-result} Test results of baselines on TREC-QA}
\end{table}

\begin{table}
\small
\centering
\begin{tabular}{|ll|ll|}
\hline  
& \bf Models & \bf MAP & \bf MRR \\
\hline
A & QA-LSTM (avg-pool) & 68.19 & 76.52 \\
B & QA-LSTM with attention & 68.96  &  78.49 \\
C & QA-LSTM/CNN & 70.61 & 81.04 \\
D & QA-LSTM/CNN with attention & \bf 71.11 & \bf 83.22 \\
E & QA-LSTM/CNN with attention & \bf 72.79 & \bf 82.40 \\
 & (LSTM hiddenvector=500) &  & \\
\hline 
\end{tabular}
\caption{\label{trec-qa-ourmodels-result} Test results of the proposed models on TREC-QA}
\vspace{-2mm}
\end{table}

\subsection{Data, metrics and baselines}

In this paper, we adopt TREC-QA, created by \cite{wangmengqiu2007} based on Text REtrieval Conference (TREC) QA track (8-13) data. 
We follow the exact approach of train/dev/test questions selection in \citep{wang2015}, in which all questions with only positive or negative answers are removed. Finally, we have 1162 training questions, 65 development questions and 68 test questions.  

Following previous work
on this task, we use Mean Average Precision
(MAP) and Mean Reciprocal Rank (MRR) as evaluation
metrics, which are calculated using the official evaluation scripts.

In Table \ref{trec-qa-baselines-result}, we list the performance of some prior work on this dataset, which can be referred to \citep{wang2015}. We implemented the Architecture II in \citep{minwei15} from scratch. 
\cite{wang2015} and \cite{minwei15} are the best baselines on MAP and MRR respectively. 

\subsection{Setup}

We keep the configurations same as those in InsuranceQA in section 4.1, except the following differences: First, we set the minibatch size as 10; Second, we set the maximum length of questions and answers as 40 instead of 200. Third, following \citep{wang2015}, We use 300-dimensional vectors that were trained and provided by word2vec
\footnote{https://code.google.com/p/word2vec/}. Finally, we use the models from the epoch with the best MAP on the validation set for training. 
Moreover, although TREC-QA dataset provided negative answer candidates for each training question, we randomly select the negative answers from all the candidate answers in the training set.

\subsection{Results}

Table \ref{trec-qa-ourmodels-result} shows the performance of the proposed models. Compared to Model (A), which is with average pooling on top of biLSTM but without attention, Model (B) with attention improves MAP by 0.7\% and MRR by approximately 2\%. The combination of CNN with QA-LSTM (Model-C) gives greater improvement on both MAP and MRR from Model (A). Model (D), which combines the ideas of Model (B) and (C), achieves the performance, competitive to the best baselines on MAP, and 2$\sim$4\% improvement on MRR compared to \citep{wang2015} and \citep{minwei15}. Finally, Model (E), which corresponds to the same model (D) but uses a LSTM hidden vector size of 500, achieves the best results for both metrics and outperforms the baselines.

\section{Conclusion}

In this paper, we study the answer selection task by employing a bidirectional-LSTM based deep learning framework. The proposed framework does not rely on feature  engineering, linguistic tools or external resources, and can be applied to any domain. We further extended the basic framework on two directions. Firstly, we combine a convolutional neural network into this framework, in order to give more composite representations for questions and answers. Secondly, we integrate a simple but efficient attention mechanism in the generation of answer embeddings according to the question. Finally, two extensions combined together. We conduct experiments using the TREC-QA dataset and the recently published InsuranceQA dataset. Our experimental 
results demonstrate that the proposed models outperform a variety of strong baselines. In the future, we would like to further evaluate the proposed approaches for different tasks, such as answer quality prediction in Community QA and recognizing textual entailment. With respect to the structural perspective, we plan to generate the attention mechanism to phrasal or sentential levels.

\bibliography{iclr2016_conference}
\bibliographystyle{iclr2016_conference}

\end{document}